\documentclass[10pt,twocolumn,letterpaper]{article}

\usepackage[pagenumbers]{cvpr}

\usepackage{graphicx}
\usepackage{amsmath}
\usepackage{amssymb}
\usepackage{booktabs}
\usepackage{wrapfig}
\usepackage{multirow}  
\usepackage{bm} \bm{}
\usepackage[accsupp]{axessibility} 
\usepackage[linesnumbered,ruled,vlined]{algorithm2e}

\usepackage[dvipsnames]{xcolor}
\definecolor{cvprblue}{rgb}{0.21,0.49,0.74}
\usepackage[pagebackref,breaklinks,colorlinks,citecolor=cvprblue]{hyperref}

\usepackage[capitalize]{cleveref}
\crefname{section}{Sec.}{Secs.}
\Crefname{section}{Section}{Sections}
\Crefname{table}{Table}{Tables}
\crefname{table}{Tab.}{Tabs.}

\def\cvprPaperID{5955} 
\def\confName{CVPR}
\def\confYear{2024}
\def\z{{\bm z}}
\def\C{{\mathcal C}}
\def\X{{\mathbf X}}
\def\Z{{\mathbf Z}}
\def\Q{{\mathbf Q}}
\def\K{{\mathbf K}}
\def\V{{\mathbf V}}
\def\muu{{\bm{\mu}}}
\begin{document}

\pagestyle{empty}  
\thispagestyle{empty} 
\title{A Video is Worth 256 Bases: Spatial-Temporal Expectation-Maximization Inversion for Zero-Shot Video Editing}
\author{Maomao Li${^{1,2}}$, Yu Li${^{2*}}$, Tianyu Yang${^{2}}$, Yunfei Liu${^{2}}$, Dongxu Yue${^{3}}$, Zhihui Lin${^{4}}$, Dong Xu${^{1*}}$ \\
${^1}$The University of Hong Kong \qquad
${^2}$International Digital Economy Academy (IDEA) \\
${^3}$Peking University \qquad
${^4}$Tsinghua University \\
{\tt\small limaomao07@connect.hku.hk \quad \{liyu,\ liuyunfei\}@idea.edu.cn} \quad {\tt\small tianyu-yang@outlook.com} \\
{\tt\small yuedongxu@stu.pku.edu.cn \quad 
lin-zhihui@outlook.com \quad dongxu@hku.hk}
}

\twocolumn[{%
\renewcommand\twocolumn[1][]{#1}%
\maketitle
\begin{center}
    \centering
    \captionsetup{type=figure}
    \vspace{-0.4cm}
    \includegraphics[width=\textwidth]{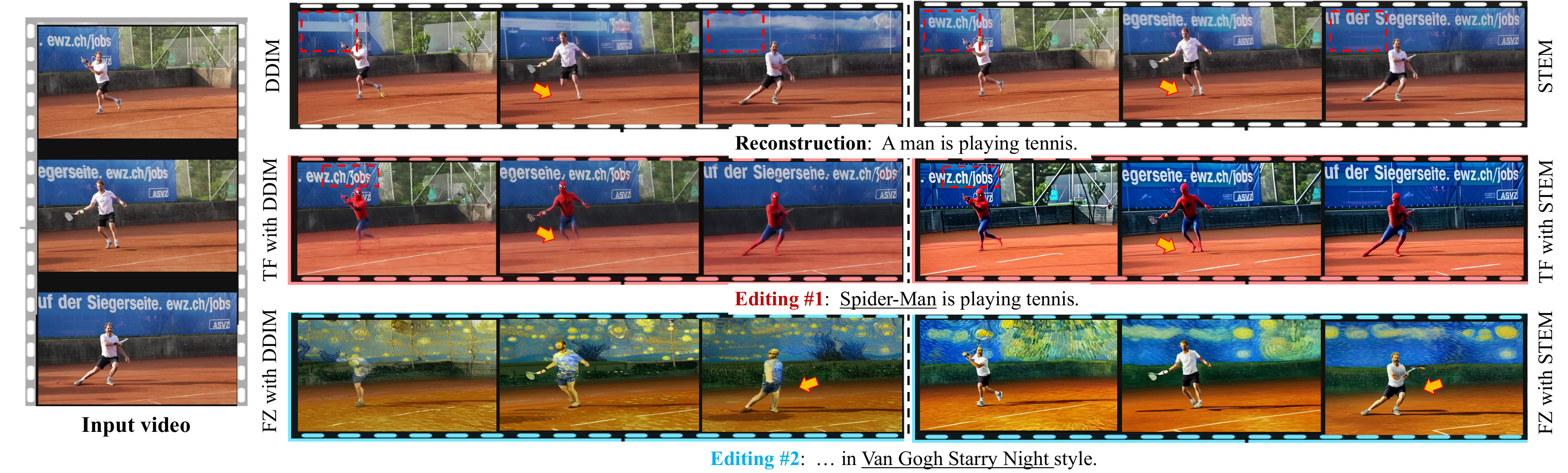}
    \vspace{-0.6cm}
       \caption{
       We propose STEM inversion as an alternative approach to zero-shot video editing, which offers several advantages over the commonly employed DDIM inversion technique. STEM inversion achieves superior temporal consistency in video reconstruction while preserving intricate details. Moreover, it seamlessly integrates with contemporary video editing methods, such as TokenFlow (TF)~\cite{tokenflow} and FateZero (FZ)~\cite{fatezero}, enhancing their editing capabilities. Best viewed with zoom-in.
       }  
    \label{fig:teaser}
\end{center}%
}]

\let\thefootnote\relax\footnotetext{*Corresponding Author}
\begin{abstract}
\vspace{-0.3cm}
This paper presents a video inversion approach for zero-shot video editing, which models the input video with low-rank representation during the inversion process. The existing video editing methods usually apply the typical 2D DDIM inversion or na\"ive spatial-temporal DDIM inversion before editing, which leverages time-varying representation for each frame to derive noisy latent. Unlike most existing approaches, we propose a \textbf{S}patial-\textbf{T}emporal \textbf{E}xpectation-\textbf{M}aximization (STEM) inversion, which formulates the dense video feature under an expectation-maximization manner and iteratively estimates a more compact basis set to represent the whole video. Each frame applies the fixed and global representation for inversion, which is more friendly for temporal consistency during reconstruction and editing. Extensive qualitative and quantitative experiments demonstrate that our STEM inversion can achieve consistent improvement on two state-of-the-art video editing methods. Project page: \url{https://stem-inv.github.io/page/}.
\end{abstract}
\vspace{-0.3cm}




\section{Introduction}
Recent years have witnessed a surge of interest in using diffusion models~\cite{ddim,ddpm} for
text-to-image (T2I) generation. There are fruitful endeavours that have been pursued, such as DALLE-2~\cite{dalle}, Imagen~\cite{imagen}, and Stable Diffusion~\cite{ldm}. Following this line, a group of methods finetune the general T2I model~\cite{ldm} for personalization usage~\cite{custom-diffusion,dreamartist,fastcomposer,dreambooth,textual-inversion,celeb}. In view of amazing results in the image domain, it is natural to leverage a pre-trained large-scale T2I model~\cite{ldm} for video editing. However, frame-wise editing inevitably brings an unacceptable flickering effect. Thus, how to effectively model 3D correspondence in a 2D model poses the main obstacle in the current video editing task.


There are three schools of research for diffusion-based text-driven video editing. 1) \cite{animatediff,magicedit}
finetune or train addition modules for the image diffusion model on massive videos to learn video motion prior. 
2) \cite{tune} tunes the image model on each input video to be edited for its temporal consistency. 
3) \cite{fatezero,pix2video,tokenflow} design frame-wise attention mechanisms to capture temporal cues without training. The last setting is more convenient for users and communities while also reaching impressive results. Therefore, we focus on the third setting in this paper.

Although DDIM inversion struggles for precise reconstruction when classifier-free guidance~\cite{classifier} is applied, most of the existing zero-shot video editing approaches only focus on making an improvement on the editing process itself to eliminate the limited editing ability brought by inaccurate inversion and reconstruction.
For example, Pix2Video~\cite{pix2video} and TokenFlow~\cite{tokenflow} use the typical DDIM image inversion to invert a video without any temporal modeling. Tune-A-Video~\cite{tune} and FateZero~\cite{fatezero} adopt a na\"ive spatial-temporal DDIM inversion, which inflates the 2D UNet of the Stable Diffusion to incorporate multiple frames for each frame inversion. To achieve acceptable complexity, each frame only explores the spatial-temporal context of two frames in total, providing rough temporal modeling.

In contrast to previous methods, this paper aims at an accurate inversion for better reconstruction and editing ability.
Specifically, we propose an efficient video inversion method, dubbed \textbf{S}patial-\textbf{T}emporal \textbf{E}xpectation-\textbf{M}aximization (STEM) inversion.
Instead of regarding all pixels in a video as the reconstruction bases~\cite{ema}, we leverage the EM~\cite{em} algorithm to find a more compact basis set (e.g., 256 bases).
Here, we treat the low-rank bases as the parameter to learn and the responsibility of each base as the latent variables in the EM algorithm. During each iteration, the E step in the proposed STEM inversion tries to calculate the expectation of latent variables (responsibility) while the M step updates the parameters (bases). The algorithm would achieve convergence by conducting E and M steps alternately for several iterations.

As for using the low-rank representation (bases) for each-frame inversion, we apply the obtained bases to generate the $\rm{Key}$ and $\rm{Value}$ embeddings in the Self-attention module, rather than applying the time-varying one frame or two frames in the sequence like previous techniques~\cite{tokenflow,pix2video,tune,fatezero}. As a result, the computational complexity of the Self-attention layer is greatly reduced.
Furthermore, we simply replace the DDIM inversion of two zero-shot video editing methods TokenFlow~\cite{tokenflow} and FateZero~\cite{fatezero} with the proposed STEM inversion. As seen in \cref{fig:teaser}, 
our method not only achieves more accurate reconstruction but also improves editing performance, \eg, the advertising banner and man's leg. To sum up, our main contributions are:
\begin{itemize}
\item 
\vspace{-0.2cm}
We propose STEM inversion for diffusion-based video editing, which reformulates a dense video under an expectation-maximization iteration manner and evaluates a more compact basis set.
\item
\vspace{-0.2cm}
Our STEM inversion can reduce the complexity and improve the reconstruction quality compared with DDIM inversion for videos even though it explores the global spatial-temporal context.
\item
\vspace{-0.2cm}
Extensive qualitative and quantitative experiments demonstrate that our STEM inversion can improve video editing performance painlessly by replacing DDIM inversion in the existing video editing pipeline.
\end{itemize}




\section{Related Work}
\noindent{\textbf{Text-dirven image generation and editing.}}
There have been many studies on image generation based on GANs~\cite{gan,stylegan}, VAE~\cite{vae,vqvae}, auto-regressive Transformers~\cite{VQVAE2,esser2021taming,gafni2022make,wu2022nuwa,yu2022scaling}, and flow~\cite{kingma2018glow}. Recently, diffusion models~\cite{ddpm,ddim} have merged as a popular choice for text-to-image (T2I) generation.
DALLE-2~\cite{dalle} and Imagen~\cite{imagen} can generate realistic images from text embeddings via large language models~\cite{clip,raffel2020exploring}, where cascaded diffusion models are used to scale up image resolution gradually.
Stable Diffusion~\cite{ldm} applies diffusion models in the latent space of powerful pre-trained auto-encoders in VQ-GAN~\cite{esser2021taming}, achieving training and inference on limited resources.

For text-driven image editing, ControlNet~\cite{controlnet} enhances large pre-trained T2I diffusion models with
task-specific image conditions, such as depth map, canny edge~\cite{canny}, and human pose. Besides, a number of studies explore text-based interfaces for content manipulation~\cite{sine,instructpix2pix}, style transfer~\cite{kwon2022diffusion,zhang2023inversion}, and generator domain adaption~\cite{kim2022diffusionclip}.


\noindent{\textbf{Text-dirven video generation and editing.}}
Text-to-video (T2V)~\cite{imagen_video,make-a-video,align} aims to generate corresponding videos based on the given prompt, which usually incorporates additional temporal domain into T2I models.
In terms of text-driven video editing, Tune-A-Video~\cite{tune} proposes to conduct one-shot video tuning for videos, while inflating 2D-UNet and replacing self-attention as sparse causal attention for temporal modeling.
Fatezero~\cite{fatezero} and Pix2Video~\cite{pix2video} apply attention maps to model temporal consistency, which can be regarded as an implicit manipulation way. 
Besides, Rerender-A-Video~\cite{rerender} and Video-ControlNet~\cite{videocontrolnet} use optical flow for frame-wise constraints.
Further, several methods~\cite{edit_a_video,video-p2p,vid2vid-zero} have been aware that the inaccurate DDIM inversion would lead to accumulated errors during the denoising process, and thus turn to Null-text inversion~\cite{null}. However, this optimization-based inversion takes 2 minutes per frame using an A100 GPU, which is unfavourable in real applications.

In contrast to previous methods, we propose to overhaul DDIM inversion into STEM inversion, where we reformulate the input video representation as a low-rank basis set by the EM algorithm, which harvests a fixed spatial-temporal context from the entire video for each frame, rather than exploring time-varying one like existing DDIM does. Besides, the row-rank representation reduces the complexity in self-attention layers.





\section{Preliminary} 
\subsection{Stable Diffusion and DDIM Inversion} 
Stable Diffusion~\cite{ldm} (SD) applies diffusion models in the latent space of powerful autoencoders, where an encoder $\mathcal{E}$ learns to encode the image $x$ into latent representations $\z=\mathcal{E}(x)$, and a decoder $\mathcal{D}$ takes the latent back to the pixel space $\mathcal{D}(\mathcal{E}(x)) \approx x$. Then, in the forward process, the model iteratively adds noise to the latent code $\z_0$, resulting in a perturbed $\z_t$:
 \begin{equation}
q(\z_t|\z_{t-1})=\mathcal{N}(\z_t;\sqrt{1-\beta_{t}}\z_{t-1},\beta_t\mathbf{I})),
\end{equation}
where $\beta_t$ denotes a hyperparameter chosen ahead of model training. In the backward process, a UNet $\epsilon_{\theta}$ is trained to predict the noise with $L_2$ objective, where the architecture consists of residual, self-attention, and cross-attention blocks.
Once trained, $\z_{0}$ can be sampled with the deterministic DDIM sampling~\cite{ddim}: 
\begin{equation}
{\z}_{t-1}\!=\!\sqrt{\frac{\alpha_{t-1}}{\alpha_{t}}}{\z}_{t} + \left(\!\sqrt{\frac{1}{\alpha_{t-1}}\!-\!1}\!-\!\sqrt{\frac{1}{\alpha_{t}}\!-\!1}\right)\!\epsilon_{\theta}({\z}_{t},t,\C),
\end{equation}
where $\alpha_{t}=\prod_{i=1}^t(1-\beta_i)$ and $\C$ is the text embedding.

DDIM Inversion aims to map an image into a known latent space (\ie, the domain of model output) before reconstruction or editing, which can be formulated as:
\begin{equation}
{\z}_{t+1}\!=\!\sqrt{\frac{\alpha_{t+1}}{\alpha_{t}}}{\z}_{t} + \!\left(\!\sqrt{\frac{1}{\alpha_{t+1}}\!-\!1}\!-\!\sqrt{\frac{1}{\alpha_{t}}\!-\!1}\right)\!\epsilon_{\theta}({\z}_{t},t,\C),
\label{ddim_inv}
\end{equation}
which is based on the assumption that the ODE process can be reversed in the limit of small steps. It works well 
for unconditional generation, while amplifies errors when classifier-free guidance~\cite{classifier} is applied. Thus, it struggles for accurate reconstruction for text-driven editing~\cite{null}.

\subsection{Expectation-Maximization Algorithm}
Formally, we denote $\X$ as the observed date and $\Z$ as the corresponding unobservable latent variables, where $\{\X,\Z\}$ represents the complete data.
As an iterative method, the expectation-maximization (EM)~\cite{em} algorithm estimates the model parameter $\theta$ by finding the maximum likelihood of the complete data.

Each iteration in the EM algorithm consists of an expectation step (E step) and a maximization step (M step). Specifically, during $r$-th iteration, 
the E step first finds the posterior distribution of the unobservable variables $\Z$ (i.e., $P(\mathbf{Z}|\mathbf{X}, \theta^{r-1})$), and then
calculates the expectation of the complete data likelihood: 
\begin{equation}
	\begin{split}
	 \mathcal{Q}(\theta, \theta^{r-1})=\sum_{\mathbf{z}}P(\mathbf{Z}|\mathbf{X}, \theta^{r-1})\ln P(\mathbf{X}, \mathbf{Z}|\theta).
	\end{split}
	\label{eq:EM_E}
\end{equation}
Then M step updates the parameter $\theta$ in $r$-th iteration by maximizing the above likelihood:
\begin{equation}
	\begin{split}
	 \theta^{r} = {\arg\max}_{\theta}\mathcal{Q}(\theta, \theta^{r-1}).
	\end{split}
	\label{eq:EM_M}
\end{equation}
The algorithm carries out the E step and M step for $R$ times to make the model converge.

\begin{figure*}[t]
  \centering
  \includegraphics[width=1.0\linewidth]{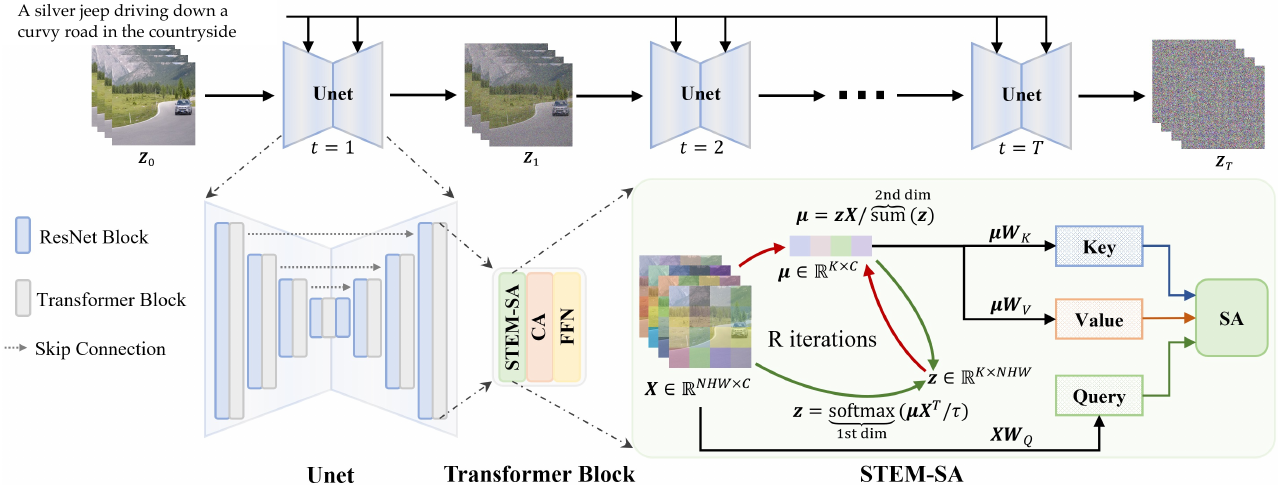}
  \vspace{-0.4cm}
  \caption{The illustration of the proposed STEM inversion method. We estimate a more compact representation (bases $\bm{\mu}$) for the input video via the EM algorithm. The ST-E step and ST-M step are executed alternately for $R$ times until convergence. The Self-attention (SA) in our STEM inversion are denoted as STEM-SA, where the $\rm{Key}$ and $\rm{Value}$ embeddings are derived by projections of the converged $\bm{\mu}$. }
  \label{stem}
\vspace{-0.3cm}
\end{figure*}

\section{Methodology}
In this section, we first revisit existing methods using DDIM inversion for video editing in Sec.~\ref{sec:3-1}. Then, the overview of the proposed STEM inversion is in Sec.~\ref{sec:3-2}. Next, we elaborate on the E step and M step in Sec.~\ref{sec:3-3} and Sec.~\ref{sec:3-4}. Finally, we introduce how to apply our STEM inversion for zero-shot video editing in Sec.~\ref{sec:3-5}.

\subsection{Using DDIM inversion For Video Editing}
\label{sec:3-1}

Given a source video sequence $\mathcal{I}=[I^1, ..., I^N]$ and its corresponding source prompt $\mathcal{P}_{src}$, a zero-shot text-driven video editing system aims to generate a new video $\mathcal{Y}=[Y^1, ..., Y^N]$ based on the target prompt $\mathcal{P}_{tgt}$ without any training or finetuning. The edited video should fully reflect the $\mathcal{P}_{tgt}$ while maintaining the temporal consistency.

To conduct video editing, the popular one-shot video editing method~\cite{tune} and zero-shot ones~\cite{fatezero,pix2video,tokenflow} first conduct DDIM inversion on each frame before the denoising process. Pix2Video~\cite{pix2video} and TokenFlow~\cite{tokenflow} use the typical 2D DDIM inversion, which inverts each frame with its own spatial context. Besides, although pre-trained SD is trained without temporal constraints, FateZero~\cite{fatezero} and Tune-A-Video~\cite{tune} inflate the 2D UNet $\epsilon_{\theta}$ and transform self-attention as spatial-temporal self-attention without changing pre-trained SD weights. Concretely, when inverting a frame, the naive spatial-temporal DDIM Inversion would exploit context from its own and another frame in the sequence.
The spatial-temporal self-attention process ${\mathrm{ATTENTION}}(\Q,\K,\V)=\mathrm{softmax}(\frac{\Q\K^{\top}}{\sqrt{d}}) \cdot \V$ for the feature $\X^n$ of $n$-th frame $I^n$ can be described as using of both $h$-th and $l$-th frame:
\begin{equation}
\Q=\X^{n}\mathbf{W}_{Q}, \K=[\X^{h},\X^{l}]\mathbf{W}_{K}, \V=[\X^{h},\X^{l}]\mathbf{W}_{V},\label{eq:selfatt}
\end{equation}
where $n, h, l \in [1,2,...,N]$, $[.]$ indicates concatenation, and $d$ is the channel number of embeddings. Fatezero employs $h=n$ and $l=\mathrm{Round}[\frac{N}{2}]$ while Tune-A-Video adopts $h=1$ and $l=n-1$.
$\mathbf{W}_{Q}, \mathbf{W}_{K}, \mathbf{W}_{V}$ are projection metrics and shared across space and time.


\subsection{The Overview of STEM Inversion}
\label{sec:3-2}
DDIM inversion provides initial noise for video reconstruction or editing, whose performance would directly influence final results.
For the typical 2D DDIM inversion, each frame only considers its own spatial context without any temporal consideration, which poses a challenge to video reconstruction and even editing.
For naive spatial-temporal DDIM inversion, each frame uses a two-frame context in the sequence to help counter the instability of temporal modeling. However, the benefit is limited.

In contrast, we argue that frames over a larger range should be considered to execute DDIM inversion.
However, using all video frames directly will bring unacceptable complexity. To deal with this, as shown in Fig.~\ref{stem}, we propose a \textbf{S}patial-\textbf{T}emporal \textbf{E}xpectation-\textbf{M}aximization (STEM) inversion method.
The insight behind this is massive intra-frame and inter-frame redundancy lie in a video, thus there is no need to treat every pixel in the video as reconstruction bases~\cite{ema}. Then, we use the EM algorithm to find a more compact basis set, which can be regarded as a general and low-rank representation for the entire video.

Formally, we denote the feature map of the input video is $\X= [\X^1,...\X^N] \in \mathbb{R}^{N \times H \times W \times C}$, where $N, C, H, W$ represent the frame number, channel number, the height, and width of the feature. Besides, we express the initial basis set as $\bm{\mu} \in \mathbb{R}^{K \times C}$, where $K$ is the number of bases.
For simplicity, we make $M=NHW$.
Since $K \ll M$, the proposed compactness is non-trivial. Specifically, we make the bases $\bm{\mu}$ as the parameters to be learned in the EM algorithm. The E step calculates the expectation of latent variables (responsibility) $\mathcal{Z}\in \mathbb{R}^{K \times M}$. The M step maximizes the likelihood of complete data to update bases $\bm{\mu}$. These two steps would be executed $R$ times before convergence.

Then, as shown in Fig.~\ref{stem}, we use the converged low-rank bases $\bm{\mu}$ for all self-attention calculations in the pre-trained SD, which is denoted as STEM-SA. Consequently, the way of calculating self-attention in Eq. (\ref{eq:selfatt}) becomes,
\begin{equation}
\begin{split}
\Q=\X^n\mathbf{W}_{Q}, \K=\bm{\mu}\mathbf{W}_K, \V=\bm{\mu}\mathbf{W}_V,
\label{eq:stem}
\end{split}
\end{equation}
where 
$\Q \in \mathbb{R}^{HW \times C}$, $\K$ and $\V$ $\in \mathbb{R}^{K \times C}$.
Note that we perform STEM-SA in all time steps and self-attention layers of UNet. For simplicity, we omit the notations of the timestep $t$ and layer $l$.

The STEM-SA mechanism in our STEM inversion would produce low-rank features and reduce complexity. For the typical 2D DDIM inversion, each self-attention layer will take $\mathcal{O}(N(HW)^2)$ computation for a N-frame video.
The complexity of the self-attention layer in naive spatial-temporal DDIM inversion is $\mathcal{O}(2N(HW)^2)$, which can be considered approximately the same as that of the typical one. In contrast, our STEM inversion reduces the complexity to $\mathcal{O}(NHWK(R+1))$. Since the iteration number $R$ is a small constant, our complexity is only $\mathcal{O}(NHWK)$. In general, we set $K \ll HW$, and the computation complexity of the proposed STEM Inversion is much less than that of DDIM inversion.
Moreover, the higher the resolution of the video, the more computation our STEM can reduce.



\subsection{ST-E Step: Responsibility Estimation}
\label{sec:3-3}

The E step in our STEM inversion conducts responsibility estimation.
This step computes the expected value
of unobservable latent variables $\mathcal{Z}=\{\mathcal{Z}_{km}\}^{K,M}_{k=1,m=1} \in \mathbb{R}^{K \times M}$, where $M=NHW$, which corresponds to the responsibility of the $k$-th basis $\bm{\mu}_k$ to $\X_m$: 
\begin{equation}
	\begin{split}
	\mathcal{Z}_{km}=\frac{\exp(\bm{\mu}_{k}\mathbf{X}^{T}_{m}/\tau)}{\sum_{j=1}^{K}\exp(\bm{\mu}_{j}\mathbf{X}^{T}_{m}/\tau)}.
	\end{split}
	\label{eq:responsibility}
\end{equation}
Here, $\tau$ is a temperature hyper-parameter controlling the shape of distribution $\mathcal{Z}$.
We also provide a more brief perspective to understand the E step in our STEM inversion. As K-means clustering~\cite{kmeans1} is a special case of EM algorithm, step E can be regarded as calculating the responsibility of $k$-th clustering center for each pixel of the video feature $\X$. That is, $\mathcal{Z}_{km}$ represents the possibility of the $m$-th pixel belonging to the $k$-th basis.
Summing up, the E step in STEM inversion can be described as:
\begin{equation}
	\begin{split}
	\mathcal{Z} = \underbrace{\mathrm{softmax}}_{\textrm{1st dim}} (\bm{\mu}\mathbf{X}^{T}/{\tau}).
	\end{split}
	\label{eq:r-e-step}
\end{equation}

\begin{algorithm}[t]
\caption{Zero-shot Video Editing Pipeline With STEM Inversion}
\label{alg:STEM}
\LinesNumbered
\newcommand{\algrule}[1][.7pt]{\par\vskip.5\baselineskip\hrule height #1\par\vskip.5\baselineskip}

\KwIn{\\
\quad feature of frames: $\X \in \mathbb{R}^{M \times C},\ M=N \times H \times W$ \\
\quad source prompt $\mathcal{P}_{src}$, target prompt $\mathcal{P}_{tgt}$
}
\KwOut{\\
\quad bases $\bm{\mu} \in \mathbb{R}^{K\times C}$ \\
\quad  the edited video $\mathcal{Y}=[Y^1, ..., Y^N]$}

\algrule

Part I: STEM Inversion

\algrule

\tcc{Setting $\tau$ as a constant.}

\For{$r=1\textrm{ to } R$}{
	\tcp{ST-\textbf{E} step, estimate responsibilities:}
     $\mathcal{Z}_{km}\gets \frac{\exp(\bm{\mu}_{k}\mathbf{X}^{T}_{m}/\tau)}{\sum_{j=1}^{K}\exp(\bm{\mu}_{j}\mathbf{X}^{T}_{m}/\tau)}$ 

	\tcp{ST-\textbf{M} step, update bases:} 
     $\bm{\mu}_{k}\gets \sum^{M}_{i=1} \frac{\mathcal{Z}_{ki}\mathbf{X}_{i}} {\sum^{M}_{m=1}{\mathcal{Z}}_{km}}$ \\
     \tcp{$m=\{1,2,...,M\}; \ k=\{1,2..,K\}$}
}
Performing STEM-SA
in DDIM-Inv (Eq.(\ref{ddim_inv})) with $\mathcal{P}_{src}$ as $\C$: \\
\quad $\Q\gets \X^n\mathbf{W}^Q$, $\K\gets \bm{\mu}\mathbf{W}^{K}$, $\V\gets \bm{\mu}\mathbf{W}^V$

\algrule

Part II: Video Editing with STEM Inversion

\algrule

\For{$t=T\textrm{ to } 1$}{
$\mathcal{Y}_{t-1}=[Y^1_{t-1}, ..., Y^N_{t-1}] \gets$   editing with spatial-temporal correspondence and $\mathcal{P}_{tgt}$.
}
\end{algorithm}

\vspace{-1mm}
\subsection{ST-M Step: Likelihood Maximization}
\label{sec:3-4}
The M step in the proposed STEM inversion aims to update the bases $\muu$ by maximizing the likelihood of the complete data (i.e., given $\X$ and estimated $\mathcal{Z}$). During the $r$-th iteration, the bases $\muu$ is updated via the following weighted summation formulation: 
\begin{equation}
 \bm{\mu}_{k} = \sum^{M}_{i=1}\frac{{\mathcal{Z}}_{ki} \mathbf{X}_{i}} {\sum^{M}_{m=1}{\mathcal{Z}}_{km}}.
 \label{eq:r-m-step}
\end{equation}
Here, the weighted summation of $\X$ encourages the bases $\bm{\mu}$ to share the same embedding space with the input feature $\X$.
To put it briefly, the M step can be described as:
\begin{equation}
    \bm{\mu} = \mathcal{Z}\mathbf{X}/\overbrace{\mathrm{sum}}^{\textrm{2nd dim}}(\mathcal{Z}).
    \label{eq:m-step}
\end{equation}
When $\tau \to 0$ in Eq.(\ref{eq:responsibility}) and Eq.(\ref{eq:r-e-step}), for any $m \in \{1,2,...,M\}$, $\mathcal{Z}_{\cdot,m}$ turns to a one-hot vector. In this way, each pixel in a video only belongs to one basis. Then, the basis would be updated by the mean of those pixels that belong to it.
In this situation, EM algorithm degenerates into K-means clustering.


\subsection{Using STEM Inversion for Video Editing}
\label{sec:3-5}
We show the pseudo algorithm of performing zero-shot video editing with our STEM inversion in Alg.~\ref{alg:STEM}. First, the video feature $\X \in \mathbb{R}^ {NHW \times C}$ are extracted from the input $\mathcal{I}=[I^1, ..., I^N]$. Next, each frame $I^n$ undergoes Eq.(\ref{ddim_inv}) to derive the noisy latent ${{\z}}_{t}^{n}$, where the STEM-SA process of $\X^n$ applies Eq.(\ref{eq:stem}).
Then, we denoise the latent with the editing process of the existing zero-shot video editing methods, such as TokenFlow~\cite{tokenflow} and Fatezero~\cite{fatezero}, where the former is based on the typical 2D DDIM inversion and the latter on the naive spatial-temporal DDIM inversion.

Besides computational complexity, our STEM inversion has a unique advantage over the typical DDIM inversion and the inflated one. Ours explores a fixed global representation rather than a time-varying one for each frame, which is more stable for reconstruction and editing.







\begin{figure*}[t]
  \centering
  \includegraphics[width=0.98\linewidth]{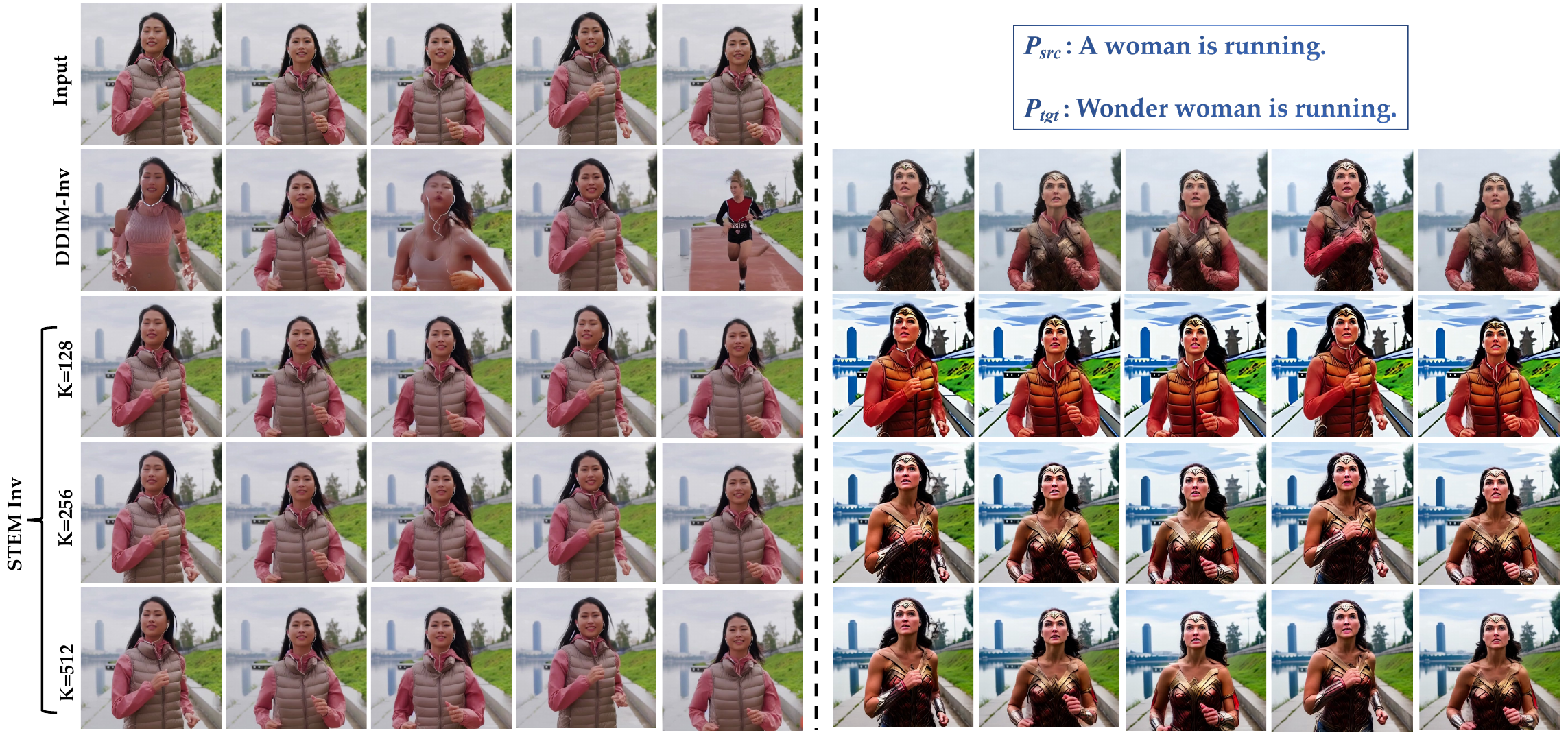}
  \vspace{-0.3cm}
  \caption{Ablation with different basis number $K$. Left: The reconstruction results of DDIM and our STEM inversion. Right: The corresponding editing results of various inversion settings, where TokenFlow editing process is used. Best viewed with zoom-in.}
  \label{fig:ablation}
  \vspace{-0.1cm}
\end{figure*} 
\begin{figure*}[t]
  \centering
  \includegraphics[width=1\linewidth]{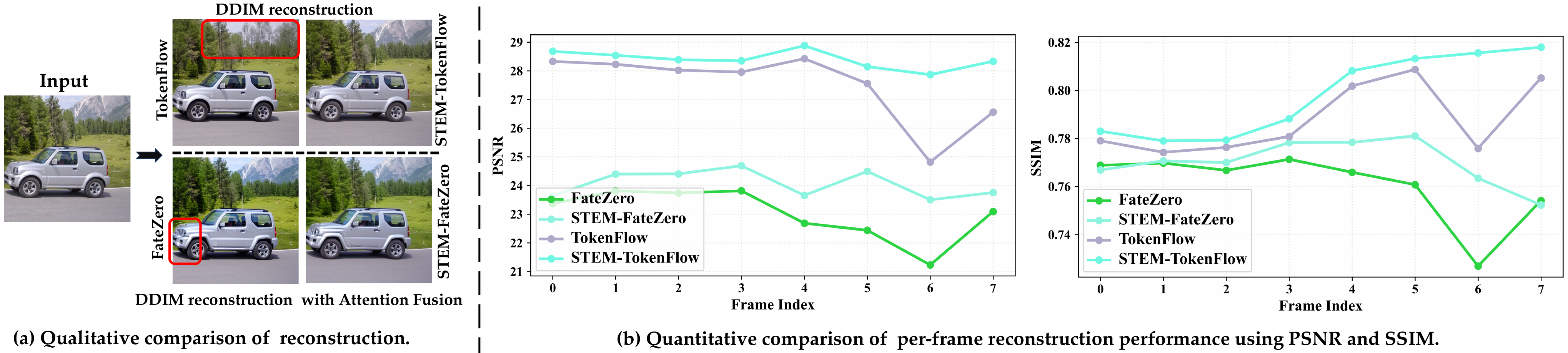}
  \vspace{-0.6cm}
  \caption{Qualitative and quantitative comparison of the reconstruction with DDIM and STEM inversion, where two reconstruction fashions are applied: (i) DDIM reconstruction, (ii) DDIM reconstruction with additional attention fusion (i.e., Fatezero~\cite{fatezero} reconstruction).}
  \label{fig:reconstruction}
   \vspace{-0.3cm}
\end{figure*}
\begin{table*}
\centering
\begin{tabular}{l|c|c|c|c|c|c}
\toprule  
\multirow{2}*{Settings}    
&\multicolumn{3}{c|}{DDIM Inversion} 
&\multicolumn{3}{c}{STEM Inversion} \\
&\textit{1-frame context}
&\textit{2-frame context}
&\textit{all-frame context}
&$K=128$ 
&{$K=256$}  
&{$K=512$}\\
\hline
Time (min) &0.80 &1.12 &5.04  &1.02 &1.50 &2.26 \\ 
\bottomrule
\end{tabular}
\vspace{-0.3cm}
\caption{Time comparison between different inversion, where the inversion steps are 50. The frame number $N$ is 48 and the spatial resolution of each frame is $640\times 360$ here. We measure them in minute (min). }
\label{tab:time}
\vspace{-0.3cm}
\end{table*}
\begin{figure*}[t]
  \centering
  \includegraphics[width=1.0\linewidth]{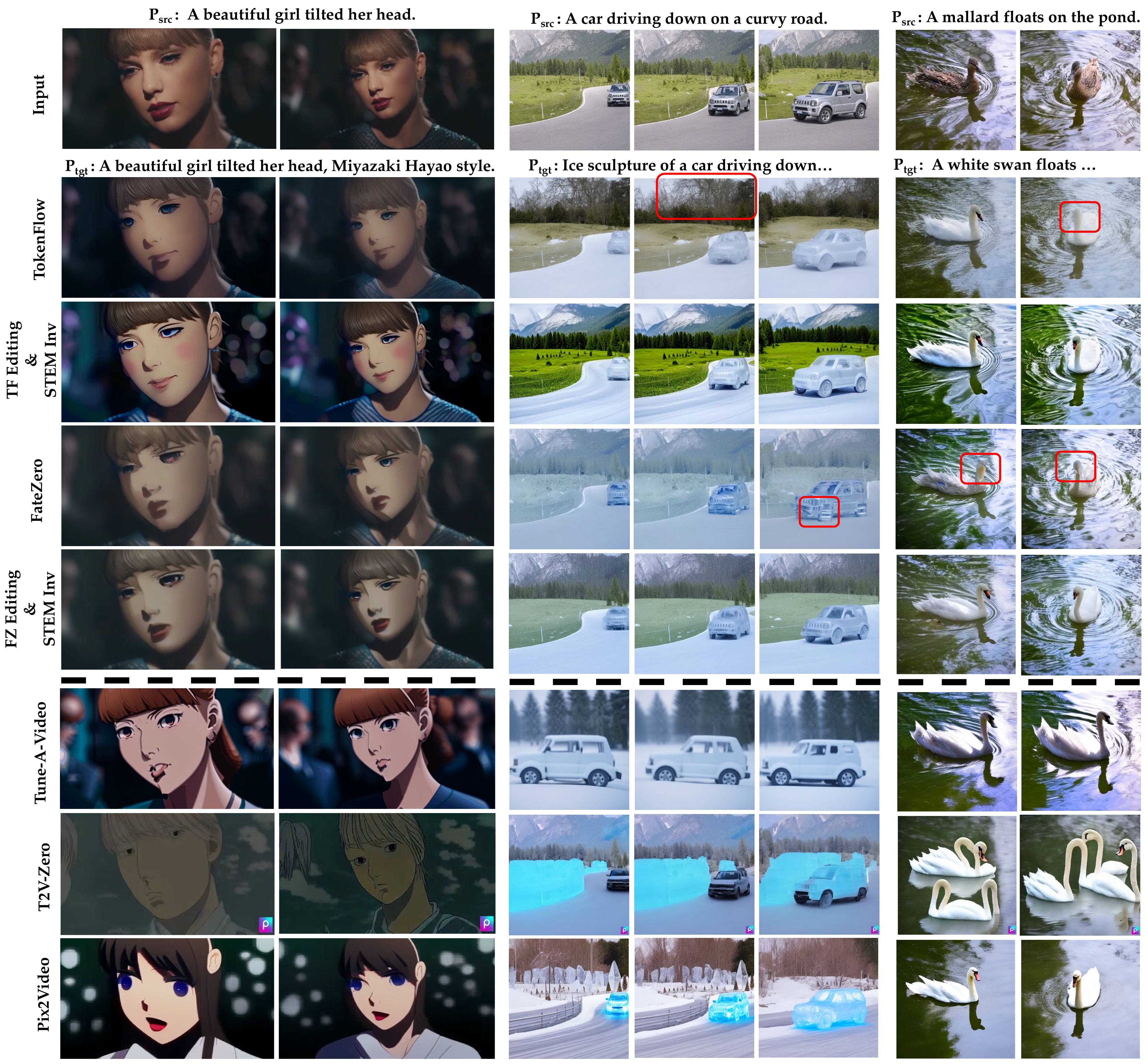}
  \vspace{-0.8cm}
  \caption{Qualitative comparison between different video editing methods. The editing scenarios here include style transfer, attribute editing, and shape editing. Best viewed with zoom-in.}
  \vspace{-0.3cm}
  \label{fig:qualitative}
\end{figure*}

\section{Experiment}
\subsection{Experimental Settings}
\noindent{\textbf{Dataset.}}
We use videos from the DAVIS dataset~\cite{davis} and the Internet for evaluation. Each video consists of $40\sim200$ frames, where the size is $512 \times 512$ or $360 \times 640$.


\noindent{\textbf{Implementation Details.}}
We use the official code of Stable Diffusion~\cite{ldm}, where the pre-trained weights are from version 1.5. In all experiments, we apply DDIM deterministic sampling with 50 steps.
Following ~\cite{fatezero,tune}, we use 50 forward steps and the classifier-free guidance of 7.5 for video editing.
Besides, we set the hyper-parameter $\tau$ in Eq.(~\ref{eq:responsibility}) as 0.05, and the iteration number $R$ as 3. 
We perform all experiments on a single NVIDIA Tesla A100 GPU.

\noindent{\textbf{Evaluation Metrics of Inversion.}}
To demonstrate the effectiveness and efficiency of the proposed STEM inversion, we use \textit{PSNR}~\cite{psnr} ($\uparrow$) and \textit{SSIM}~\cite{ssim}($\uparrow$) to evaluate the quality of each frame in the reconstructed video.

\noindent{\textbf{Evaluation Metrics of Edited Videos.}}
Following~\cite{tokenflow,pix2video}, our first metric \textit{CLIP score}($\uparrow$) measures
edit fidelity, which calculates the similarity of CLIP embedding~\cite{clip} of each edited frame and the target prompt. The second metric \textit{warp error} ($\downarrow$) reflects temporal consistency. We first estimate optical flow~\cite{raft} of the input video, which is used to warp the edited frames. Then, we compute the average MSE between warped frames and the target ones.


\vspace{-0.1cm}
\subsection{Ablation Study}
\vspace{-0.1cm}
To investigate the influence of basis number $K$, we report the reconstruction and editing performance when $K$ is 128, 256, and 512 respectively. Here, the video resolution is $512\times 512$, and the frame number $N$ is 120. As shown in Fig.\ref{fig:ablation}, 
$K=256$ achieves a similar reconstruction and editing performance with $K=512$. It shows that 256 bases are sufficient to represent the entire video. However, when $K=128$, although our STEM inversion still derives competitive reconstruction, the editing ability is weakened compared with $K=256$. Thus, we use $K=256$ as default.

\vspace{-0.05cm}
\subsection{Inversion Comparison}
\vspace{-0.05cm}
\noindent{\textbf{Qualitative comparison.}}
We apply an 8-frame video to evaluate performance, where we consider two reconstruction fashions: (i) the typical DDIM reconstruction, used via TokenFlow~\cite{tokenflow}; (ii) additionally leveraging attention fusion during DDIM reconstruction, used via FateZero~\cite{fatezero}. Fig.~\ref{fig:reconstruction} shows STEM inversion consistently boosts the performance of DDIM reconstruction and the one with attention fusion. Especially, the distant mountains can be accurately reconstructed via ours which DDIM inversion failed. See more reconstruction comparisons in our supplement.

\begin{figure}[t]
  \centering  \includegraphics[width=1.0\linewidth]{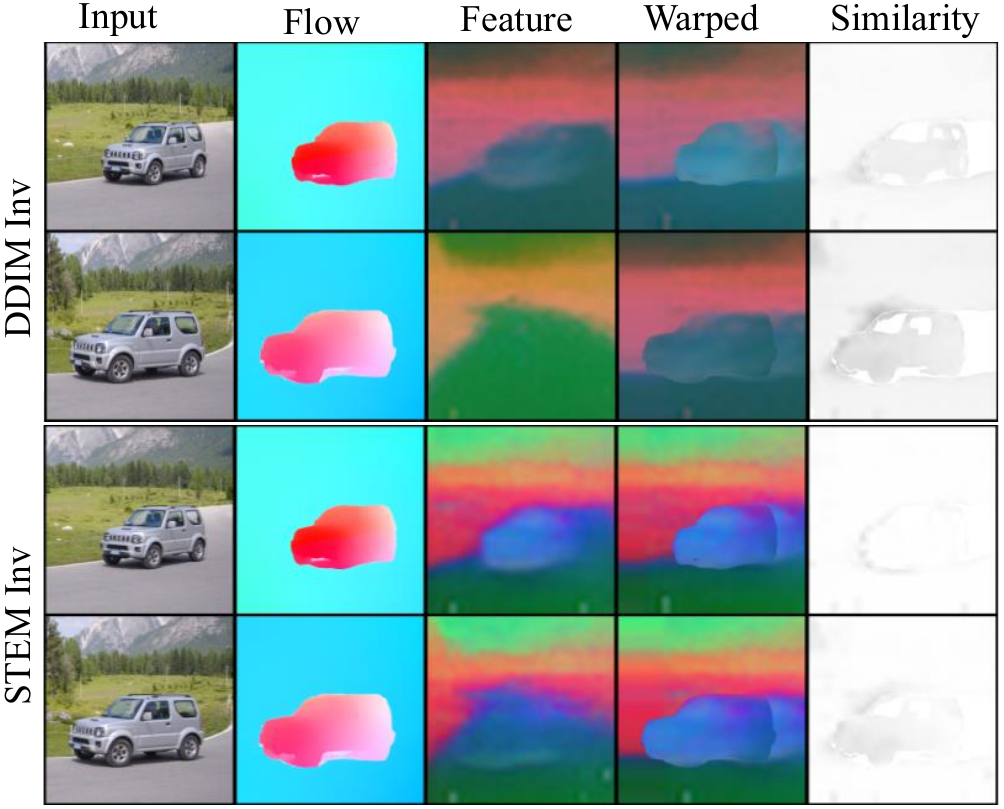}
  \vspace{-0.6cm}
  \caption{Visualization of different inversions. 
  We first estimate optical flow~\cite{raft} of the input video. Then, we apply PCA to the features of last SA layer from the decoder with different inversion. Next, 
  we use the optical flow to warp the features of former-frame to obtain the warped features in the 4-th column. Last, we show the cosine similarity of features from 3-rd and 4-th column.
  }
  \label{fig:vis}
    \vspace{-0.3cm}
\end{figure}
\noindent{\textbf{Quantitative comparison.}}
We give a quantitative evaluation of reconstruction quality with different inversions. In Fig.~\ref{fig:reconstruction}, we report \textit{PSNR} and \textit{SSIM} of each reconstructed frame, where the typical DDIM reconstruction (TokenFlow and STEM-TokenFlow) and the one with attention fusion~\cite{fatezero} (FateZero and STEM-FateZero) are both listed. We find our STEM inversion always achieves better performance than DDIM inversion for video reconstruction.

\noindent{\textbf{Feature visualization.}} In Fig.~\ref{fig:vis}, we first estimate the optical flow~\cite{raft} of the input video. Then, we apply PCA on the output features of the last SA layer from the UNet decoder. 
The 4-th column shows the feature visualization when we use optical flow to warp the former-frame features.
Last, we give the cosine similarity of the warped features and the target ones. Here, the brighter, the better temporal consistency can be achieved from the perspective of optical flow.
The cosine distance here is similar to the \textit{Warp error}. Specifically, \textit{Warp error} calculates MSE, which is suitable for RGB difference while cosine distance is more reasonable for inversion features with multiple channels.

\noindent{\textbf{Time Cost.}}
We measure time cost for various inversions, where a 48-frame video with $640\times 360$ size is applied. 
As seen in Table.\ref{tab:time},  the typical DDIM inversion using 1-frame context for each frame costs the smallest time 0.8 min. 
The naive spatial-temporal DDIM inversion exploring a 2-frame context would increase the time cost by 40\%, yielding 1.12 min. We also provide a more radical inflated DDIM version for reference, which uses all-frame context and needs 5.04 min.
In contrast, our STEM ($K=128$ and $K=256$) not only harvests the spatial context from the entire video but also achieves a similar time cost with naive spatial-temporal DDIM inversion using a 2-frame context. More comparison of our STEM inversion with the typical 2D DDIM inversion and the inflated one can be found in the supplement.

One may wonder why STEM inversion can reduce computational complexity, but it still takes more time than the typical DDIM inversion. The reason is the iterations of  EM algorithm are serial, which is unfavourable for time cost to some extent.
Both the radical inflated DDIM inversion and ours can explore the global context from the entire video, but our time cost is very advantageous compared with it. 


\subsection{Comparison of Video Editing}
\vspace{-0.1cm}
\noindent{\textbf{Qualitative Comparison.}}
We first give the qualitative comparison of editing results when using DDIM and our STEM inversion respectively. From Fig.~\ref{fig:qualitative}, we find that STEM inversion can improve the editing performance of both TokenFlow and FateZero.
Moreover, FateZero struggles to perform shape editing with Stable Diffusion (see Mallard example). Instead, it usually uses the pre-trained model~\cite{tune} for shape editing.
Fortunately, by inserting our STEM inversion, we can empower FateZero with shape editing capabilities.
Besides, we compare our method with current state-of-the-art zero-shot video editing: Tune-A-Video~\cite{tune},  Pix2Video~\cite{pix2video}, and Text2Video~\cite{text2video}, where ours exceed them on both edit fidelity and temporal consistency. We give more examples in the supplement.


\begin{table}
\centering
\small
\setlength{\tabcolsep}{1.3mm}{} 
\begin{tabular}{l|c|c}
\toprule  
Method 
&\textit{CLIP score $\uparrow$}
&\textit{Warp error (1e-3) ($\downarrow$)} \\
\hline
Tune-A-Video~\cite{tune} &0.28&32.3 \\
Pix2Video~\cite{pix2video} &\textbf{0.32}&6.2 \\
Text2Video-Zero~\cite{text2video} &0.31&22.7 \\
FateZero~\cite{fatezero} &0.29&7.2 \\
TokenFlow~\cite{tokenflow} &0.31&4.9 \\
\hline  
Ours (STEM-FateZero)  &0.30&4.3  \\
Ours (STEM-TokenFlow)  &0.31& \textbf{3.5} \\
\bottomrule
\end{tabular}
\vspace{-0.3cm}
\caption{Qualitative comparison with other video editing methods.}
\label{tab:quantitative}
\vspace{-0.2cm}
\end{table}
\noindent{\textbf{Quantitative Comparison.}}
The quantitative comparison is in Table.~\ref{tab:quantitative}. 
We achieve a competitive CLIP score and the lowest warp error. Moreover, compared with FateZero (or TokenFlow),  our STEM-Fatezero (or STEM-TokenFlow) can always yield better performance, demonstrating the superiority of our STEM inversion.

\begin{table}
\small
\centering
\setlength{\tabcolsep}{1.3mm}{} 
\begin{tabular}{l|c|c}
\toprule  
Method 
&\textit{edit fidelity}
&\textit{temporal consistency} \\
\hline
FateZero~\cite{fatezero} &17\% & 33\% \\
Ours (STEM-FateZero) &\textbf{83\%} & \textbf{67\%}
\\
\hline  
TokenFlow~\cite{tokenflow} &25\% &33\% 
 \\
Ours (STEM-TokenFlow)  &\textbf{75\%} &\textbf{67\%}  \\
\bottomrule
\end{tabular}
\vspace{-0.3cm}
\caption{User study results (\%), where we show the averaged selection percentages of each method.}
\label{tab:user_study}
\vspace{-0.3cm}
\end{table}
\noindent{\textbf{User Study.}}
We provide a user study for FateZero \textit{vs} STEM-FateZero and TokenFlow \textit{vs} STEM-TokenFlow.
Corresponding to two key objectives of video editing, we ask users to choose:
\textit{a) the one with the higher edit fidelity, and 
b) the one with the better temporal consistency.}
For each user, we randomly sample 12 videos. 
We report the selected ratios in Table.~\ref{tab:user_study}. Those editings cooperating with our STEM inversion always earned the higher preference.

\section{Conclusion}
In this paper, we present a STEM video inversion method. Concretely, like EM algorithm, we evaluate a more compact basis set by performing E step and M step iteratively. Considering our STEM inversion does not require any finetuning, it can be
seamlessly integrated into other diffusion-based video editing methods that utilize DDIM inversion as initial noise. With its clear technical advantages, we hope that our STEM inversion can serve as a new inversion method for video editing for future contributions.

While our work shows promising results, there are still some limitations. For example, it chose 256 bases as the low-rank representation of the whole video. However, for an extremely long or high-resolution video, 256 bases would be not enough to obtain sufficient editing ability. An adaptive number of bases is more reasonable. We will leave it as future work.
\clearpage

\clearpage
{\small
\bibliographystyle{ieee_fullname}
\bibliography{main}
}

\appendix


\section{High Correspondence in Diffusion Feature} 
\label{A}
\begin{figure}[t]
  \centering  \includegraphics[width=1.0\linewidth]{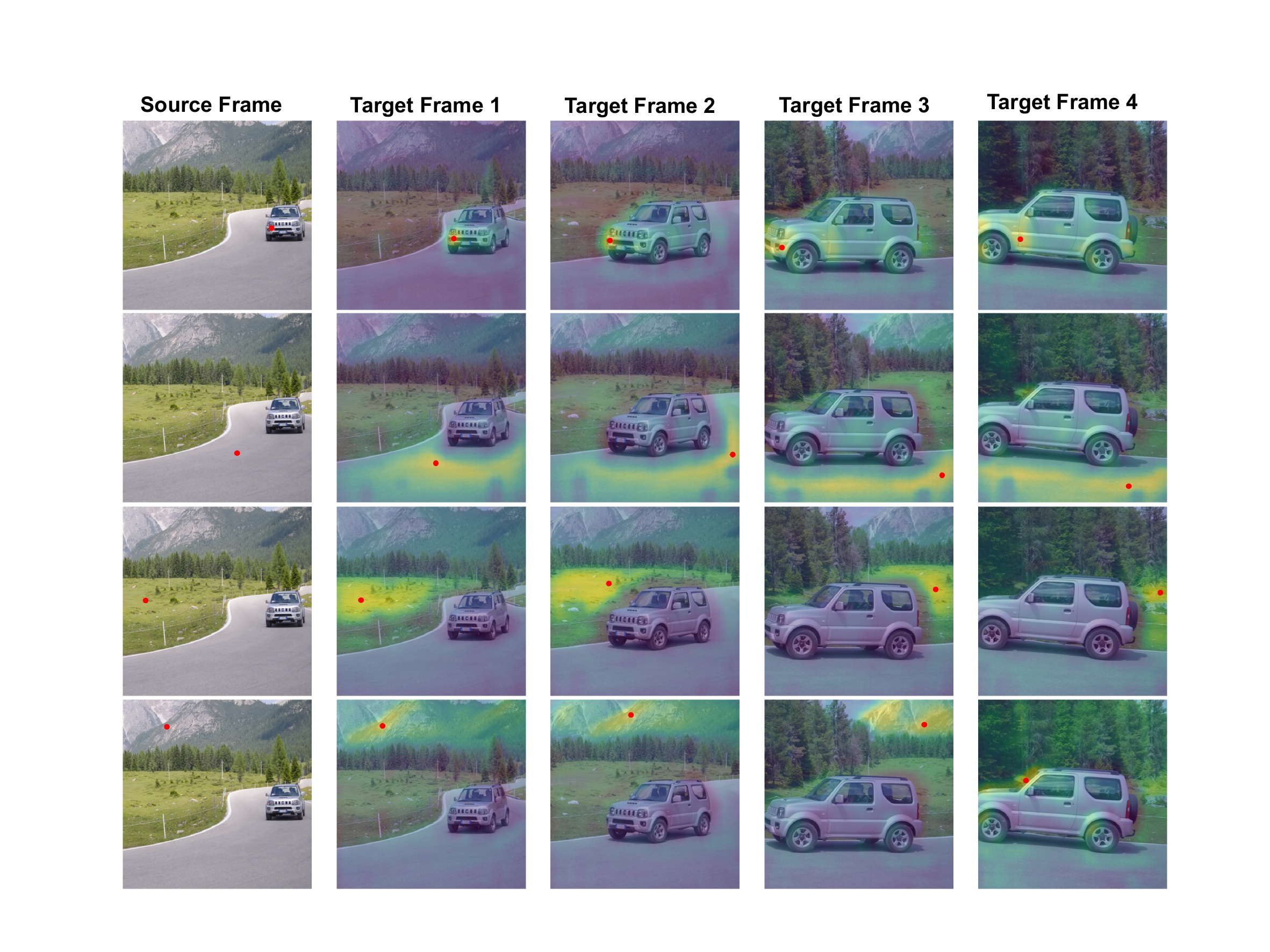}
  \caption{Given a different source pixel, the best matching pixel from the target frames can be predicted via diffusion features.}
  \label{fig:dift}
\end{figure}

The previous work ~\cite{emergent} has proven that {\textit{correspondence emerges in image diffusion models (e.g., StableDiffusion) without any explicit supervision}}, where diffusion features can be used to find matching pixel locations in two images by a simple nearest neighbor lookup. Fig~\ref{fig:dift} shows correspondences between video frames using diffusion features. Since there are high correlations in video diffusion features, we can use the EM algorithm to identify the low-rank representation for the entire video.

\let\thefootnote\relax\footnotetext{*Corresponding Author}

\section{Inversion Comparison}
\label{B}

\subsection{DDIM Inversion Using All-frame Context}
\begin{figure*}[t]
  \centering  \includegraphics[width=1.0\linewidth]{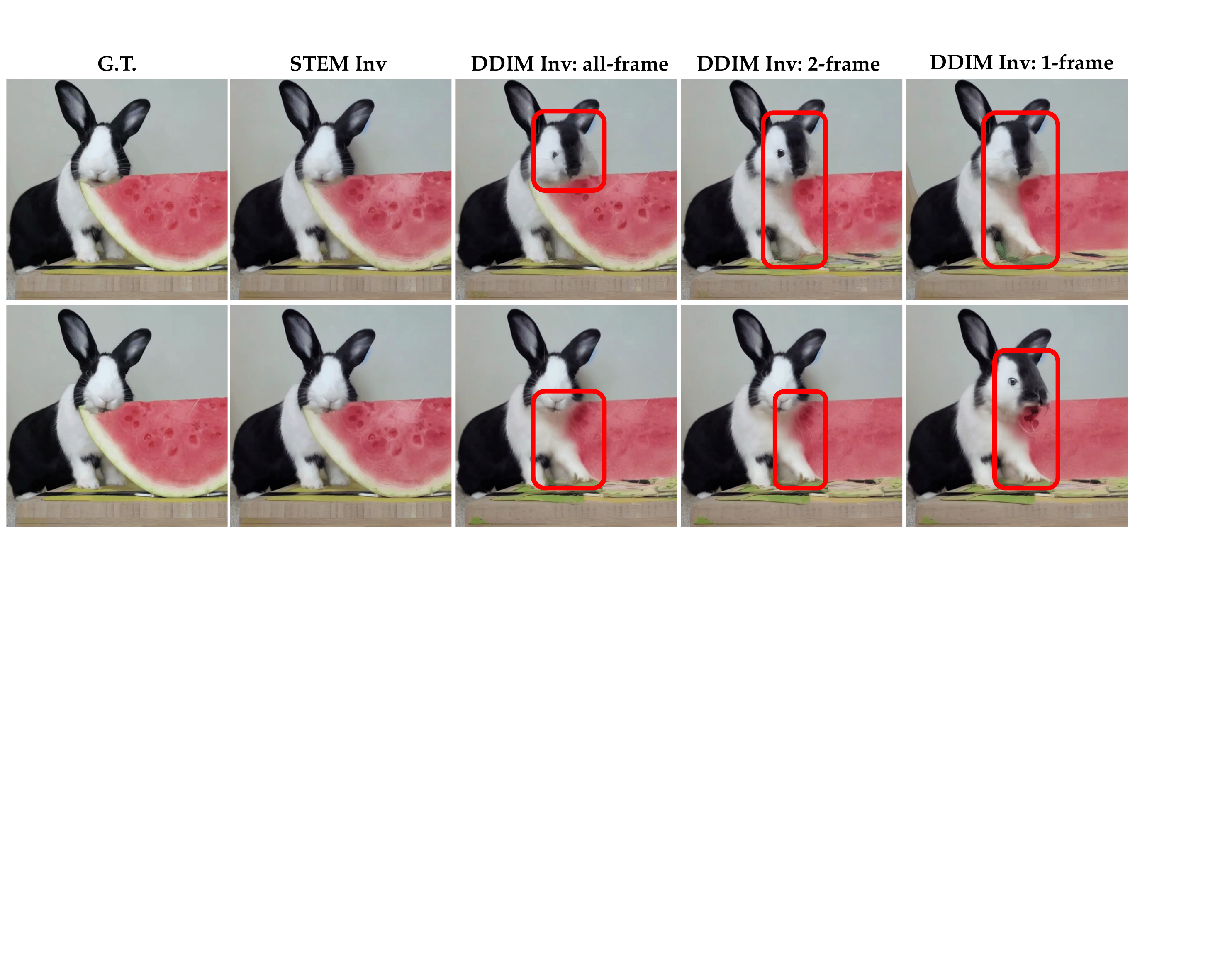}
\caption{Reconstruction results with DDIM inversion and the proposed STEM inversion, respectively.}
  \label{fig:all-frame}
\end{figure*}

Recall that DDIM inversion in existing video editing methods~\cite{tune,fatezero,tokenflow} usually exploits 1-frame or 2-frame context to invert each frame. Thus, we design a more radical inflated DDIM inversion that uses all-frame context as a reference in Table 1 of our paper. Here, we use the typical DDIM reconstruction method to provide a video reconstruction comparison in Fig.~\ref{fig:all-frame}, where both our STEM inversion and the radical inflated DDIM one can explore context from the entire video, while the resource-consuming latter yields inferior performance. The quantitative comparison in Tab.~\ref{tab:all-frame} also supports these findings, where we record the average PSNR and SSIM of 5 reconstruction videos. 
We argue that the underlying reason is those redundant or abnormal features can be effectively removed by evaluating low-rank representations.

\begin{figure}[t]
  \centering  \includegraphics[width=1.0\linewidth]{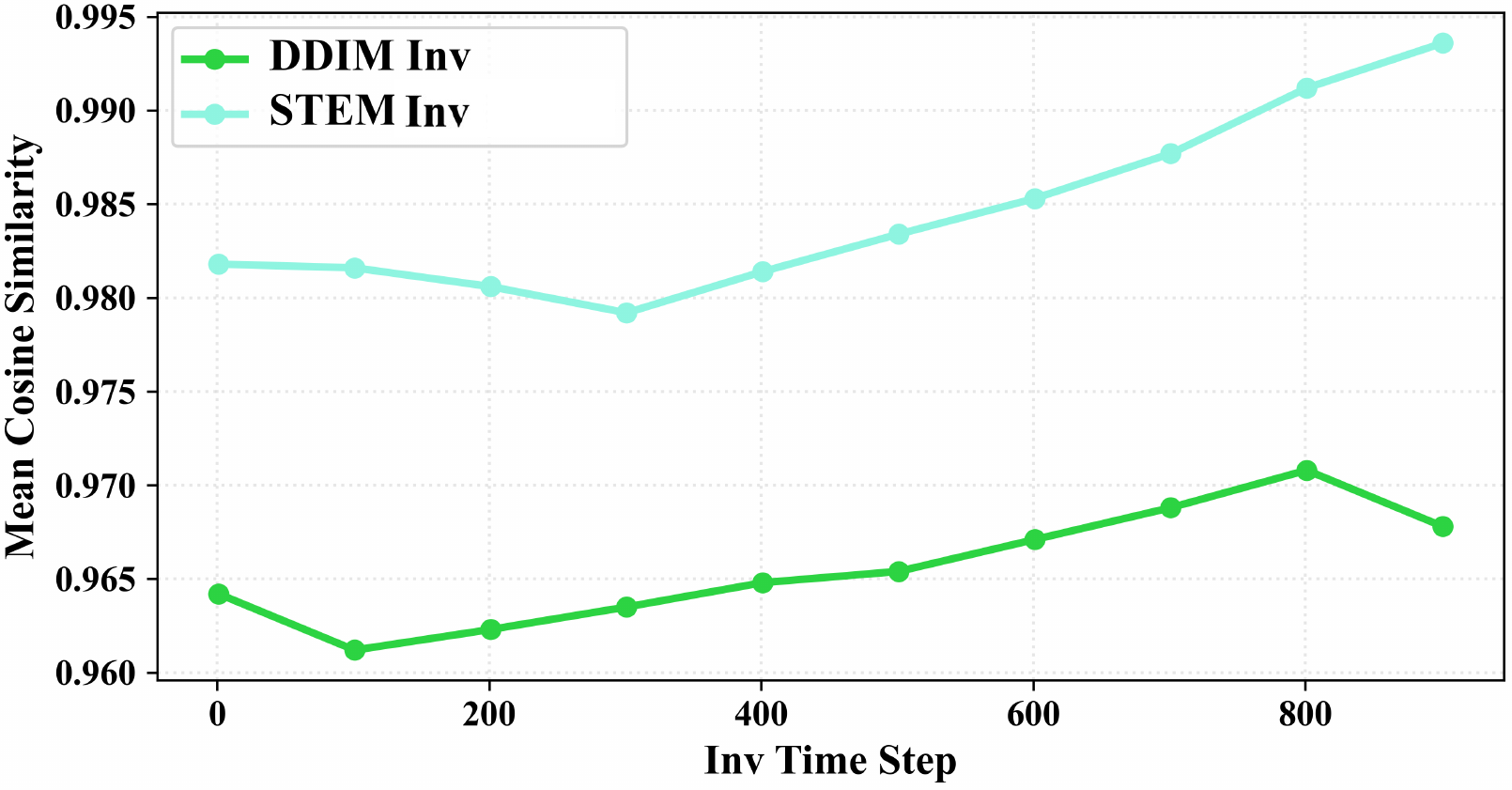}
  \caption{The mean cosine similarity between the warpped features from the former frame and the target features during various inversion steps. The more similar, the better.}
  \label{fig:simi}
\end{figure}


\subsection{Feature Similarity in Different Forward Steps}
Recall that in Fig. 6 of our main paper, we first use optical flow to warp the former-frame features and obtain the warped feature. Then, we calculate the cosine similarity between the wrapped feature from the former frame and the current frame feature. The more similar, the reconstructed video is more coherent in time dimension.

In this supplement, we provide the mean cosine similarity across different time steps $t$ in Fig.~\ref{fig:simi}. The higher similarity
indicates that our STEM inversion can achieve better temporal consistency from the perspective of optical flow.

\subsection{STEM Inversion with Various Video Lengths and Video Resolutions} 
We provide average PSNR and SSIM between 5 reconstruction videos and ground-truth ones in Tab.~\ref{tab:sm-length} and Tab.~\ref{tab:sm-resolution}. For Tab.~\ref{tab:sm-length}, we sample original videos evenly whose size is $512^2$, and form 8, 16, 32, 64, and 128-frame videos separately. For Tab.~\ref{tab:sm-resolution}, the frame number is 24.
Our STEM inversion achieves the best reconstruction when the frame number is 16 and the resolution is $1024^2$.

\begin{table}
\centering 
\begin{tabular}{l|c|c|c|c|c}
\toprule  
Length
&{8}
&{16}
&{32} 
&{64}
&{128}\\
\hline
PSNR &32.702  &\textbf{33.279}  &33.197   &33.058    &32.751  \\
\hline
SSIM  &0.9836 &\textbf{0.9854}  &\textbf{0.9854}    &0.9851   &0.9843 \\
\bottomrule
\end{tabular}
\caption{Results of STEM inversion with
various video lengths.}
\label{tab:sm-length}
\end{table}
\begin{table}
\centering
\begin{tabular}{l|c|c|c|c}
\toprule  
Size 
&{$384^2$}
&{$512^2$}
&{$768^2$} 
&{$1024^2$}\\
\hline
PSNR &29.558 &33.767   & 34.813    &\textbf{35.668}  \\
\hline
SSIM  &0.9743 &0.9847  &0.9901 &\textbf{0.9902}  \\
\bottomrule
\end{tabular}
\caption{Results of STEM inversion with various video resolutions.}
\label{tab:sm-resolution}
\end{table}

\begin{table*}
\centering
\begin{tabular}{c|c|c|c|c|c|c|c}
\toprule 
\multicolumn{2}{c|}{DDIM Inv (1-frame)} & 
\multicolumn{2}{c|}{DDIM Inv (2-frame)}  & 
\multicolumn{2}{c|}{DDIM Inv (all-frame)} & 
\multicolumn{2}{c}{STEM Inv}
\\
PSNR &SSIM 
&PSNR &SSIM
&PSNR &SSIM 
&PSNR &SSIM \\
\hline
24.122 &0.8137 &25.967 &0.8595 &26.464 &0.8700  &\textbf{31.572}  &\textbf{0.9606} \\
\bottomrule
\end{tabular}
\caption{Qualitative comparison between different inversion. Here, ``1-frame", ``2-frame", and ``all-frame" refer to the context frames considered during single-frame inversion calculation for DDIM inversion.}
\label{tab:all-frame}
\end{table*}

\begin{figure*}[t]
  \centering
\includegraphics[width=1\linewidth]{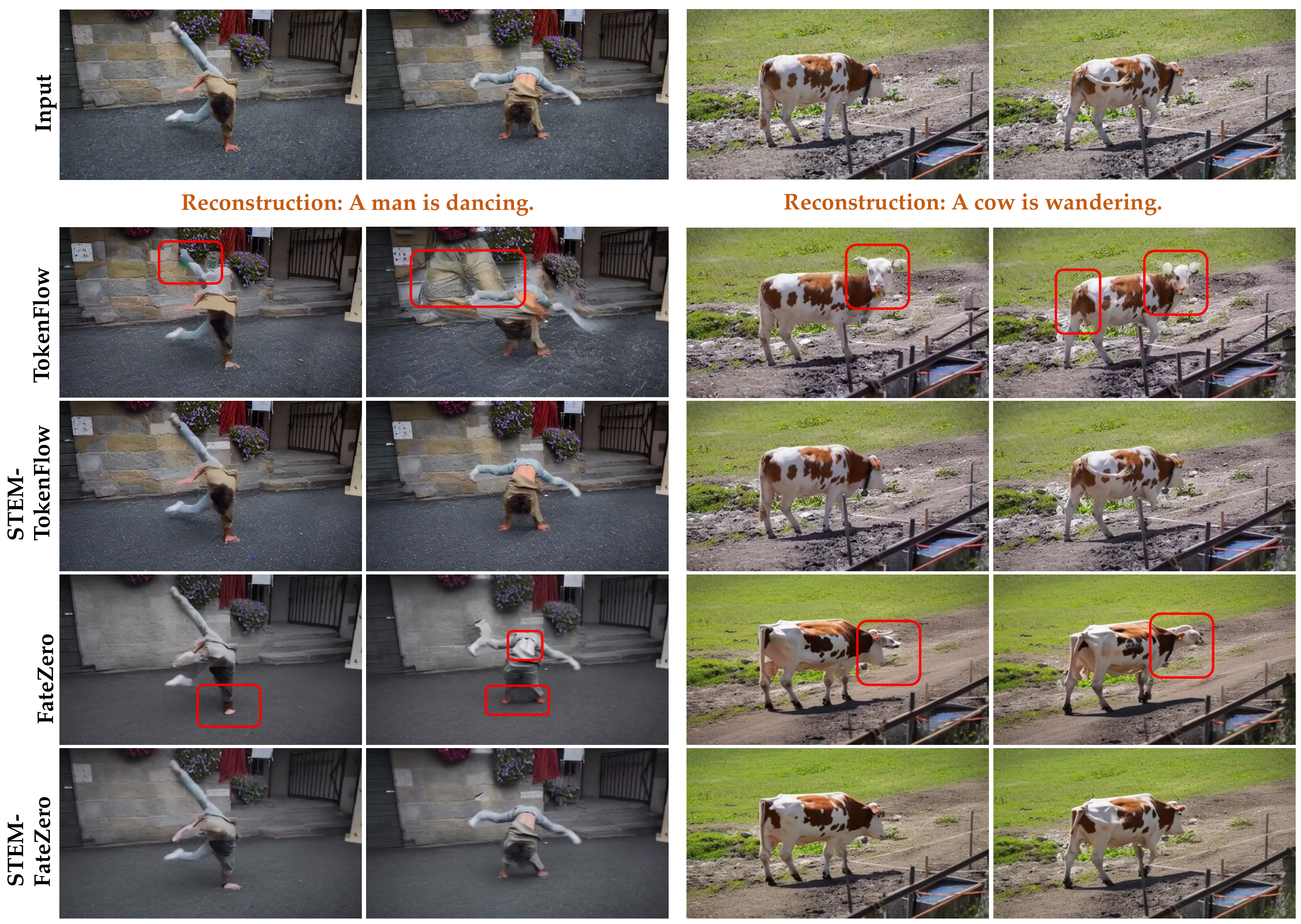}
  \caption{Qualitative comparison of the reconstruction with DDIM and STEM inversion, where two reconstruction fashions are applied: (i) DDIM reconstruction (i.e., TokenFlow~\cite{tokenflow} reconstruction)), (ii) DDIM reconstruction with additional attention fusion (i.e., Fatezero~\cite{fatezero} reconstruction).}
  \label{fig:sm-reconstruction}
\end{figure*}

\subsection{More Reconstruction Comparison}
We provide the reconstruction comparison of DDIM inversion and STEM inversion in Fig.~\ref{fig:sm-reconstruction} and our project page. Since Fatezero~\cite{fatezero} stores the intermediate self-attention maps and cross-attention maps at each timestep $t$, it is memory-consuming and cannot perform video editing over 20 frames on a single A100 GPU. On our project page, we sample FateZero results at a proper rate.

As seen in Fig.~\ref{fig:sm-reconstruction},
two reconstruction fashions are applied for DDIM and STEM inversion separately: (i) the typical DDIM reconstruction (used by TokenFlow~\cite{tokenflow}), (ii) DDIM reconstruction with extra attention fusion (used by FateZero~\cite{fatezero}).
The proposed STEM inversion always delivers better reconstruction than DDIM inversion, especially under the typical DDIM reconstruction fashion.
Such a benefit is derived from our STEM inversion modelling global and fixed context for each frame while DDIM inversion explores a time-varying and limited spatial-temporal context.

\section{More Comparison of Various Text-driven Zero-shot Video Editing}
\label{C}

\begin{figure*}[t]
  \centering \includegraphics[width=1.0\linewidth]{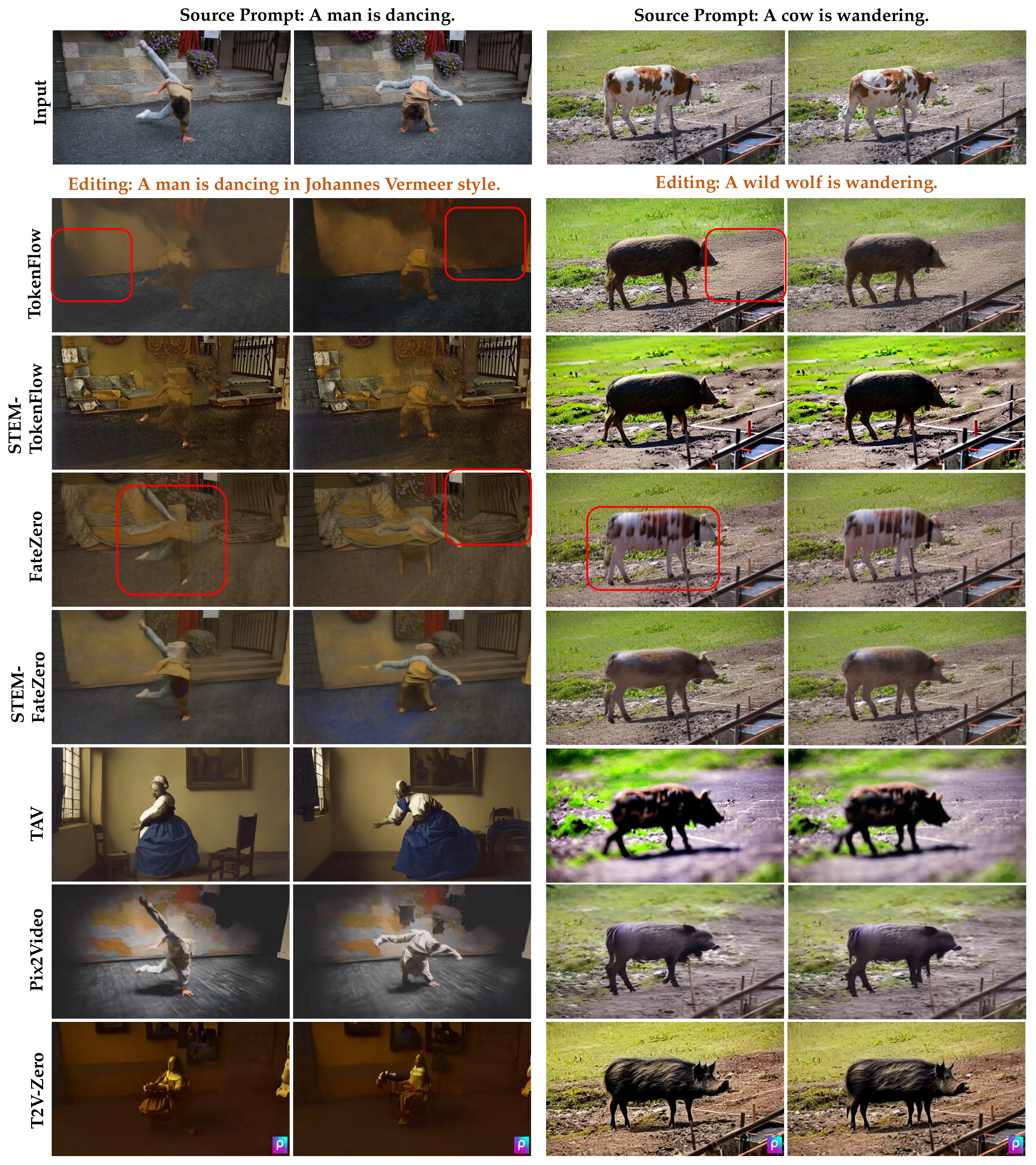}  \caption{Qualitative comparison between different text-driven video editing methods. Our STEM-inversion can consistently improve the editing performance of TokenFlow~\cite{tokenflow} and FateZero~\cite{fatezero}. Best viewed with zoom-in.}
  \label{fig:sm-edit}
\end{figure*}

To prove the efficiency of our STEM inversion, 
we compare our STEM-TokenFlow and STEM-FateZero with the current state-of-the-art video editing methods in Fig.~\ref{fig:sm-edit} and our project page.
Specifically, although T2V-Zero~\cite{text2video} can perceive the style and subject to be edited, it always deviates greatly from the original video and cannot maintain a satisfying temporal consistency.
Besides, Tune-A-Video~\cite{tune} needs to perform training on the video before editing while its performance is inferior to ours.
Moreover, it is difficult for Pix2Video~\cite{pix2video}
to maintain the background. Please see the second from last row of Fig.~\ref{fig:sm-edit}.

Note that FateZero~\cite{fatezero} struggles to conduct shape editing (see ``cow'' $\to$ ``boar''). Our STEM inversion is able to endow shape-editing ability to FateZero, which also demonstrates the superiority of our method.
Besides, by replacing DDIM inversion with the proposed STEM inversion, TokenFlow also yields more high-quality video editing results. As seen in Fig.~\ref{fig:sm-edit},
our STEM-TokenFlow has better editing fidelity when transferring the video to \textit{Johannes Vermeer style}.

\section{More Implementation Details}
\label{D}
The attention fusion ratio in 
FateZero~\cite{fatezero} is a hyperparameter controlling the editing effect. Specifically, it fuses both cross-attention and self-attention at DDIM time step $t\in[0.5T,T]$. However, we experimentally discovered a small fusion ratio for cross attention time step is better when using our STEM inversion for Fatezero editing. Concretely, we adopt the cross attention time step ratio as $t\in[0.2T, T]$, while the same ratio $t\in[0.5T, T]$ for self-attention. The possible underlying reason is that our STEM inversion is more sensitive to capturing the semantics from the target prompt than the DDIM one. Thus, a smaller cross-attention fusion ratio is sufficient under the FateZero editing scenario. Besides, in terms of TokenFlow editing, we use identical hyper-parameters when replacing DDIM inversion with our STEM inversion.

\end{document}